\pdfoutput=1
\documentclass[11pt]{article}

\usepackage[final]{acl}

\usepackage{times}
\usepackage{latexsym}
\usepackage{tikz}
\providecommand{\keywords}[1]{\textbf{\textit{Keywords—}} #1}
\usepackage{lipsum}
\usepackage[labelfont=bf]{caption}
\usepackage{tabularray}
\usepackage{tcolorbox}
\usepackage{verbatim}
\usepackage{fancyvrb}
\usepackage{academicons}

\usepackage{colortbl}
\PassOptionsToPackage{table}{xcolor}
\usepackage{geometry}
\usepackage[T1]{fontenc}
\usepackage[utf8]{inputenc}

\usepackage{microtype}

\usepackage{inconsolata}

\usepackage{hyperref}
\usepackage{graphicx}

\title{FinStat2SQL: A Text2SQL Pipeline for Financial Statement Analysis}

\author{
 \textbf{Quang Hung Nguyen\textsuperscript{1}},
 \textbf{Phuong Anh Trinh\textsuperscript{1}},
 \textbf{Phan Quoc Hung Mai\textsuperscript{1}},
 \textbf{Tuan Phong Trinh\textsuperscript{1}}
\\
\\
 \textsuperscript{1} College of Technology, National Economics University, Vietnam
\\
 \small{
   \textbf{Correspondence:} \href{mailto:quanghung20gg@gmail.com}{quanghung20gg@gmail.com}
 }
}

\begin{document}
\title{FinStat2SQL: A Text2SQL Pipeline for Financial Statement Analysis}
\makeatletter
\let\@oldmaketitle\@maketitle%
\renewcommand{\@maketitle}{\@oldmaketitle%
\includegraphics[width=\textwidth, clip]{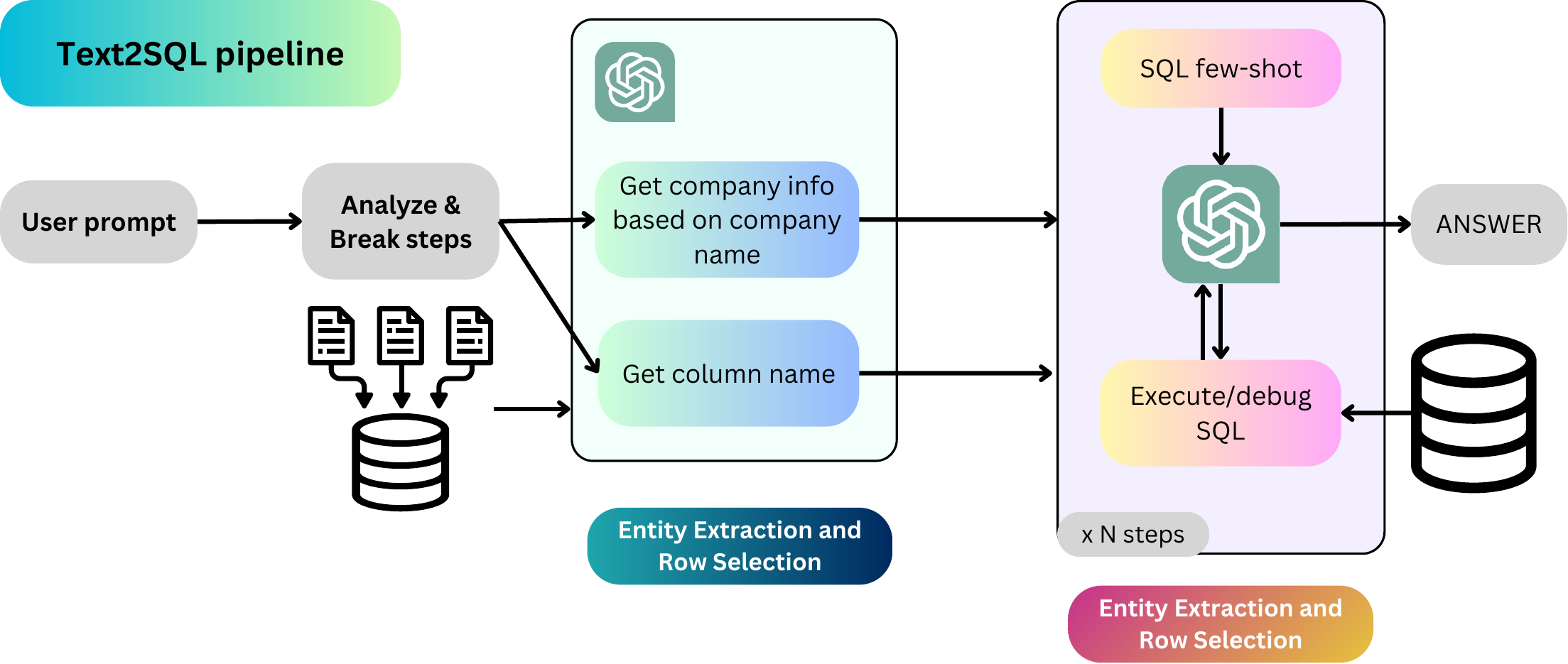}
\vspace{-0.6cm}
\captionof{figure}{Diagram of the FinStat2SQL Pipeline: The language model first parses and analyzes the user's prompt, retrieves relevant company database information, and then iteratively executes and debugs SQL queries until a correct result is obtained, which is returned as the final answer. More information can be found in \textbf{Sec. \ref{sec:Approach}}.}
\label{fig:text2sqlpipeline}
\vspace{0.2cm}
}

\makeatother

\maketitle
% \begin{figure*}[h]
%     \centering
%     \includegraphics[width=\textwidth, clip]{pipeline.pdf}
%     \caption{Diagram of the FinStat2SQL Pipeline: The language model first parses and analyzes the user's prompt, retrieves relevant company database information, and then iteratively executes and debugs SQL queries until a correct result is obtained, which is returned as the final answer.}
% \label{fig:text2sqlpipe}

% \end{figure*}
\begin{abstract}
Despite the advancements of large language models, text2sql still faces many challenges, particularly with complex and domain-specific queries. In finance, database designs and financial reporting layouts vary widely between financial entities and countries, making text2sql even more challenging. We present FinStat2SQL, a lightweight text2sql pipeline enabling natural language queries over financial statements. Tailored to local standards like VAS, it combines large and small language models in a multi-agent setup for entity extraction, SQL generation, and self-correction. We build a domain-specific database and evaluate models on a synthetic QA dataset. A fine-tuned 7B model achieves 61.33\% accuracy with sub-4-second response times on consumer hardware, outperforming GPT-4o-mini. FinStat2SQL offers a scalable, cost-efficient solution for financial analysis, making AI-powered querying accessible to Vietnamese enterprises.
\end{abstract}


\section{Introduction}

Generating accurate SQL from users’ natural language questions (text2sql) is a long-standing challenge \cite{hong2024next}, especially in the scenario that both the database system and user queries are growing more complex. For years, many attempts have been carried out \cite{katsogiannis2023survey}, introducing more sophisticated methods: from deep neural networks \cite{lei2020re, li2023resdsql} to Generative Language Models \cite{NEURIPS2023_83fc8fab,pourreza2023din}. The results of Generative Language Models are very promising, however, many challenges persist when dealing with domain-specific tasks, which require further studies and refinement into each task. Database systems continue to grow rapidly and support for more languages becomes increasingly necessary, the potential applications of text2sql are broad—particularly in financial database systems, where structured querying via natural language can greatly enhance accessibility and efficiency. Applications of text2sql are wide, for example, integrating into systems such as financial reporting automation, customer support chatbots, regulatory monitoring, as well as healthcare record querying, academic research databases, supply chain tracking, HR analytics \cite{singh2025survey}.

Financial reports are essential for communicating a company’s financial health and informing the decisions of investors, regulators, and analysts. However, manual analysis is often time-consuming, prone to human error, and inefficient for professionals who require fast, accurate insights from large volumes of financial data. While many companies globally follow the International Financial Reporting Standards (IFRS) issued by the International Accounting Standards Board (IASB) \cite{wagenhofer2009global}, national adaptations can vary significantly \cite{albu2014global}. For instance, \citet{nguyen2014measurement} highlights that Vietnam Accounting Standards (VAS) only partially converge with IFRS, resulting in structural and interpretational differences. This diversity adds complexity to querying financial indicators across jurisdictions. To address these challenges, many text2sql techniques have gained traction as a promising solution for automating and streamlining financial analysis \cite{li2024codes, zhang2024finsql}. 

In this research, we design a pipeline for generating SQL commands from financial statistics ( \textbf{Figure~\ref{fig:text2sqlpipeline}}), and evaluate the SQL query generation capabilities of various language models within the context of the Vietnamese Companies Financial Statement System. Our study encompasses both Large Language Models (LLMs) and Small Language Models (SLMs), assessing their effectiveness in translating natural language questions into executable SQL queries. In addition, we benchmark multiple training strategies for Supervised Fine-Tuning (SFT) of SLMs to determine which approaches yield the best performance for domain-specific tasks. Lastly, we explore the practical integration of these text2sql models into a financial analysis system, demonstrating their potential to enhance data accessibility, query accuracy, and response efficiency.

\section{Literature Review}

In this section, we outline the challenges of querying financial indexes, provide a brief overview of recent advances in text2sql methods, and present key techniques for tuning language models.

\subsection{Problem in Financial Standards convergence}
Financial statements are important and efficient for users, whether they are accountants, executive boards, government officials, or equity holders. Most accounting systems worldwide follow IFRS in preparing financial statements \cite{ifrsjuri}. IFRS \cite{ifrs15} sets principles for recognizing, measuring, and disclosing financial elements. IFRS 15 governs revenue recognition through a five-step model, IFRS 16 addresses lease accounting by capitalizing most leases, and IFRS 9 regulates financial instruments with an expected credit loss model, promoting transparency and comparability. \citet{de2016review} revealed that IFRS adoption offers substantial benefits, including greater transparency, reduced capital costs, enhanced cross-border investment, improved financial comparability, and increased foreign analyst coverage.

However, some countries do not fully apply IFRS. In the case of Vietnam, this country applies a national standard, called VAS, which is not completely identical to IFRS. While IFRS is principle-based, VAS is rule-based. Although  \citet{phan2014perceptions} and \citet{phan2018influences} discussed many Vietnam intentions to converge with IFRS, \citet{nguyen2014measurement} further showed that VAS and IFRS only achieve mid-level convergence. When these conflicts still exist, it means that financial reporting systems are still recorded differently, causing many difficulties in querying. We endeavor to address this problem by researching and training automatic queries using language models, and because cases like Vietnam are not unique.

\subsection{Text2SQL} 
\noindent \textbf{Data.} In recent years, numerous datasets have been introduced to support the development of text2sql systems. The Spider dataset \cite{yu-etal-2018-spider}, for instance, spans 138 diverse domains to challenge cross-domain generalization. WikiSQL \cite{zhong2017wikisql} offers a large-scale resource with over 24,000 tables derived from Wikipedia. Other datasets like Squall \cite{shi-etal-2020-squall} and KaggleDBQA \cite{lee-etal-2021-kaggledbqa} further explore model generalization on unseen schemas. In contrast, several domain-specific datasets have been curated for more focused evaluation, including those based on Yelp and IMDB reviews \cite{imdb}, the Advising dataset SEDE \cite{hazoom-etal-2021-sede}, and datasets targeting the Restaurants \cite{tang:ecml01-restaurant} and Academic \cite{academic} domains. These datasets aim to assess model performance within narrow domains, often prioritizing precision over generalization. However, these datasets still do not fully satisfy the demands of the industry, which are often more complex and challenging \cite{zhang2024finsql}.

\vspace{0.20 cm}

\noindent\textbf{Text2SQL methods.} Prior to LLMs, text2sql relied on fine-tuning encoder-decoder models. To improve representation, graph-relational neural networks, such as LGESQL\cite{cao2021lgesql} or RATSQL \cite{wang2019rat}, were used to capture structural relationships among query tokens, tables, and columns. With the rise of large language models like T5 and LLaMA, fine-tuning methods have significantly improved text2sql performance on benchmarks like Spider. Recent advances include Graphix \cite{gan2021-20} for multi-hop reasoning, Picard \cite{iyer2017-30} for constrained decoding, and RESDSQL \cite{gan2021-19} is the current state-of-the-art fine-tuning method on the Spider leaderboard. To improve SQL generation, MAC-SQL \cite{wang2023mac} and E-SQL \cite{caferouglu2024esql} decompose queries into sub-steps, aiding data understanding and reducing logical errors. However, this multi-step approach increases processing time due to added computational overhead.

\vspace{0.20 cm}

\noindent\textbf{ Text2SQL in Finance.} Despite its potential application, very little research has been conducted on the application of text2sql in finance. FinSQL \cite{zhang2024finsql} is a model-agnostic LLM-based text2sql framework tailored for financial analysis, addressing challenges like wide tables and limited domain-specific datasets. Alongside the introduction of the BULL benchmark—sourced from real-world financial databases—FinSQL achieves state-of-the-art performance with efficient fine-tuning and prompt design, showing up to 36.64\% improvement in few-shot cross-database settings. \citet{kumar2024booksql} introduces BookSQL, a large-scale text2sql dataset focused on the accounting and financial domain, featuring 100k NL-SQL pairs and databases with over 1 million records. It highlights the challenges current models face in this domain, emphasizing the need for specialized approaches to support non-technical users in querying accounting databases.

% \subsection{Language model tuning method}
% \vspace{-0.2 cm}
% \noindent\textbf{Parameter-Efficient Fine-Tuning.} Parameter-Efficient Fine-Tuning (PEFT) techniques, such as Adapter \cite{houlsby2019adapter}, Prompt Tuning \cite{lester2021prompt}, Prefix-Tuning \cite{li2021prefix}, and LoRA \cite{hu2022lora}, modify only a small portion of the parameters in large pre-trained language models. These methods can achieve performance similar to full fine-tuning while requiring far fewer computational resources. Typically, less than 1\% of the model’s weights are adjusted, allowing the model to quickly adapt to new downstream tasks using just a single GPU. This efficiency makes PEFT particularly attractive for scenarios with limited hardware availability or when rapid iteration and deployment are needed.

% ADD DPO and KTO

% \vspace{0.2 cm}

% \noindent\textbf{Chain of thought.} Chain of Thought (CoT) \cite{cot} is a technique that allows LLMs to approach complex reasoning tasks in a manner similar to human thinking before producing the final result. In this approach, LLMs generate a step-by-step reasoning process before arriving at the final answer, which significantly improves their performance on reasoning challenges. CoT is well known, and is applied in many language models, such as Google Palm-540B \cite{chowdhery2023palm}, where it dramatically increases accuracy from 17.9\% to 56.5\% in arithmetic reasoning. 

\section{Preliminaries}
\subsection{Database Construction}
In this research, we constructed a financial database using data from 2016 to Q3 2024, covering 200 major Vietnamese companies listed on VN30, HNX30, and others across diverse industries. Data was collected from the FiinPro website, processed from Excel files, and stored in a structured SQL database. The schema consists of four components: company details, financial statements, financial ratios, and vector databases for entity matching. To resolve inconsistencies in Vietnam’s three financial statement formats (banks, corporations, securities), we implemented a universal mapping table to unify account codes and prevent data conflicts.

To evaluate model performance under varying constraints, we created two database checkpoints: a training database with 102 companies and limited data, and a comprehensive testing database simulating full-scale deployment with 200 companies and additional accounts and fields. This dual setup allows us to assess the pipeline’s scalability and robustness, ensuring it can generalize across smaller and larger datasets while adapting to structural and chronological changes in financial data.

To enable efficient financial statement analysis, we adopt a STAR schema database, structuring key financial figures in central fact tables linked to descriptive dimensions (e.g., company, industry, category). This design supports fast, flexible queries over historical data, improves performance by reducing complex joins, and ensures scalability and consistency through standardized mappings \cite{iqbal2019star}. The database, as mentioned before, is often large and complicated. For reference, see \textbf{Appendix \ref{erd}} for our database construction.

\subsection{Synthetic QA Dataset}

\begin{table}[t]
\centering
\begin{tabular}{l c}
\hline
\textbf{Task Type} & \textbf{Count} \\
\hline
Financial \& Accounting Conversations & 1,800 \\
Entity (Account) Extraction Tasks     & 4,000 \\
SQL Task Conversations                & 11,000 \\
\hline
\end{tabular}
\caption{Synthetic training dataset composition.}
\label{tab:synthetic_data}
\end{table}

To train the language model effectively for financial analysis, we created a large-scale synthetic question-answer dataset focused on exploring, interpreting, and analyzing financial statements. This dataset was generated through an automated pipeline designed to cover a broad range of tasks, including fundamental, technical, and comparative financial analysis. We used Gemini 2.0 Flash Thinking Experimental 01-21 and GPT-4o mini as the main generators for this synthetic data. The example of example golden query and result is represented at \textbf{Appendix \ref{exgoldenquery}}.

To ensure quality, we applied a rigorous selection process using the LLM-as-a-Judge framework \cite{zheng2023llmjudge}, where each generated pair was reviewed by multiple independent evaluators. Only high-quality, contextually relevant, and analytically sound pairs were retained for fine-tuning, while suboptimal examples were repurposed for alignment training. The final synthetic dataset is illustrated in\textbf{ Table \ref{tab:synthetic_data}}.

\subsection{Evaluating Dataset}

To comprehensively evaluate the FinStat2SQL pipeline in real-world scenarios, we curated a dataset of around 300 financial analysis questions. These were sourced from actual financial reports by stock exchanges, brokerage firms, and investment analysts, covering company performance, market sectors, and industry trends. Although many were originally presented in chart format, we reformulated them into SQL-based tasks to align with our system’s capabilities.

To further strengthen the evaluation, we augmented the dataset with multiple-choice questions (MCQs). For each financial task, we created one to five MCQs to assess the model’s reasoning, contextual understanding, and retrieval accuracy. This approach ensures a more thorough and realistic test of the system’s ability to interpret and analyze structured financial data.

\section{Approach}
\label{sec:Approach}
In this section, we introduce FinStat2SQL, a pipeline that converts noisy financial queries into accurate SQL using entity extraction, code generation, and self-correction (\textbf{Figure \ref{fig:text2sqlpipeline}}).

\subsection{Text2SQL Pipeline}

\begin{figure}[h]
    \centering
    \includegraphics[width=\columnwidth, clip]{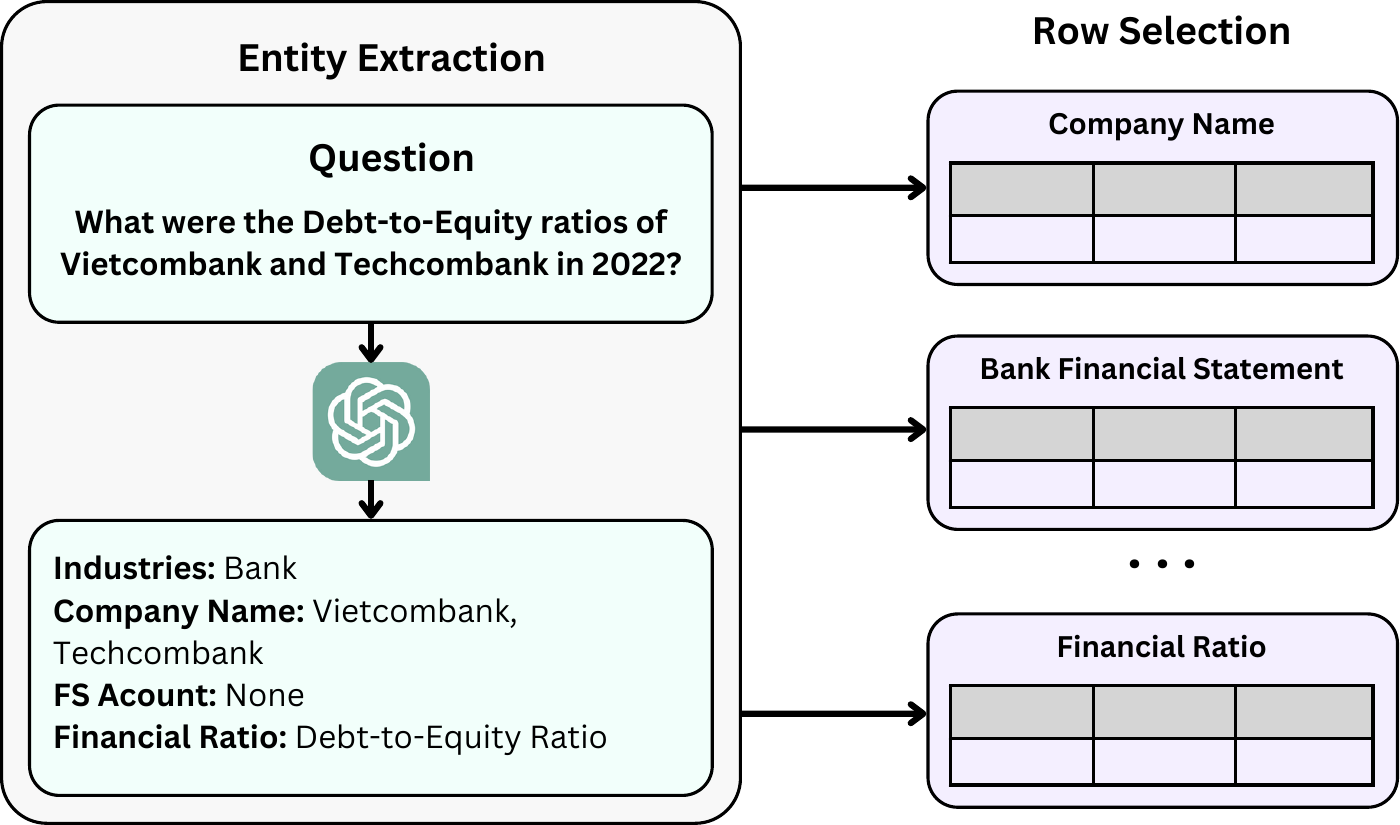}
    \caption{ Diagram of Entity Extraction and Row Selection process}
    \label{fig:err}
\end{figure}

\textbf{Entity Extraction.} The Entity Extraction step uses LLMs to identify key elements in user queries for generating accurate SQL, focusing on four fields: Industry, Company Name, Financial Statement Account, and Financial Ratio. To achieve this, the agent utilizes a prompt-based approach with LM (see \textbf{Appendices \ref{entityprompt}} and \textbf{\ref{schemadesc}} for the prompts). This prompt-based approach minimizes ambiguity, ensures precise parsing, and allows the system to infer additional relevant metrics, offering more flexibility and accuracy than traditional NER methods (see \textbf{Figure~\ref{fig:err}}).

\textbf{Row Selection.} After entity extraction, matching candidates are retrieved from vector databases, with a selection mechanism used to narrow down relevant results, especially when entities are ambiguous. The selected candidates are then passed to the Code-Generation LLM to determine the best match based on context.

\begin{figure*}[h]
    \centering
    \includegraphics[width=\textwidth, clip]{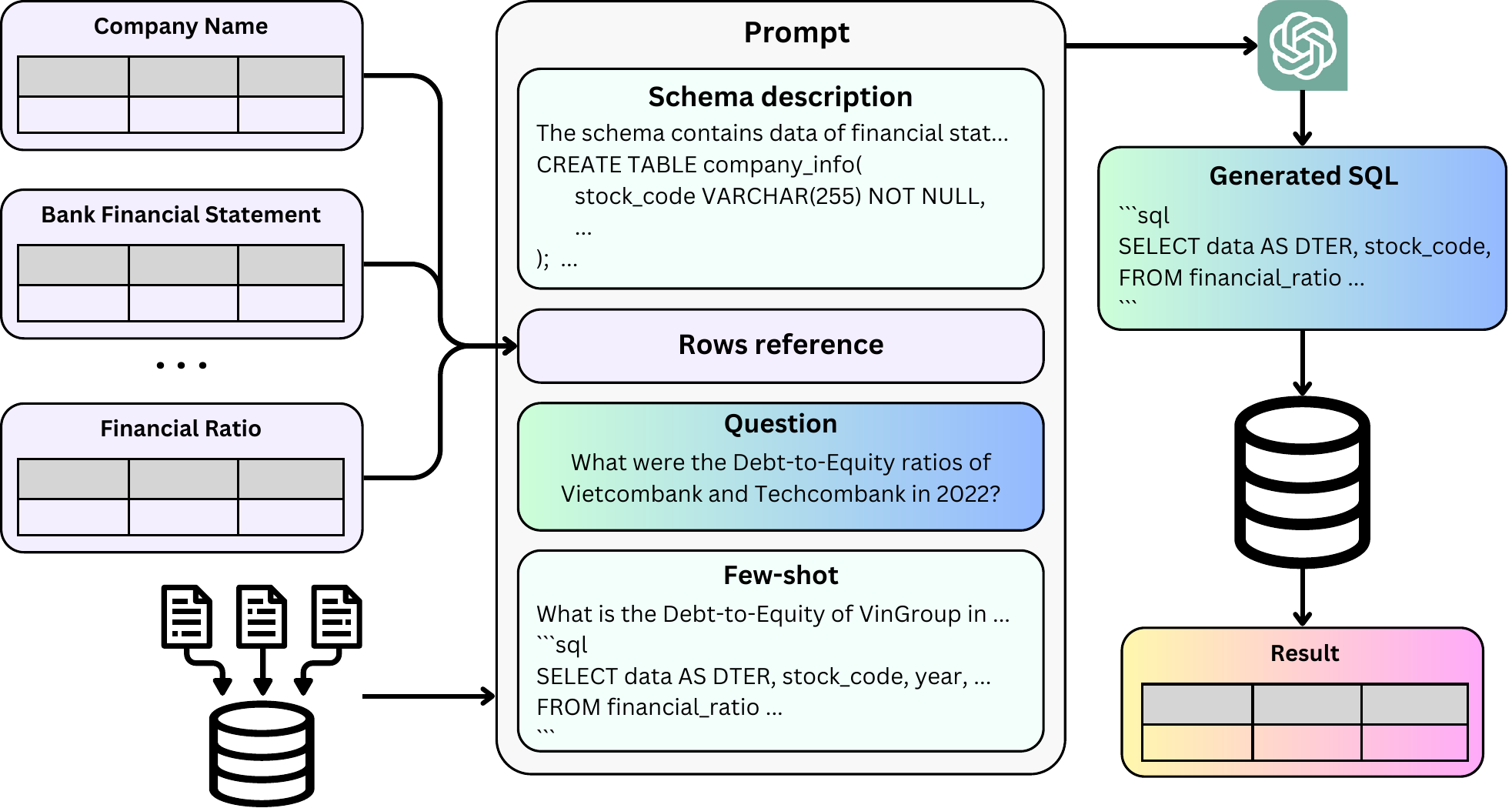}
    \caption{SQL generation process}
    \label{fig:sqlgen}
\end{figure*}

\textbf{Similarity Search.} Full-text search is insufficient for extracting relevant entities in specialized domains like financial statements under VAS due to semantic nuances, such as the difference between "Net Income" (IFRS) and "Profit After Tax" (VAS), which it cannot reconcile. Unlike full-text search, we leverage similarity search using vector databases to map entities based on their semantic meaning, ensuring accurate alignment even with different phrasing. By incorporating similarity search in our RAG system, we bridge the gap between commonly trained LLMs and VAS-specific terminology, enhancing entity matching and improving the accuracy of query results in specialized domains.

From this paragraph, we describe three SQL Generation process, see \textbf{Figure \ref{fig:sqlgen}} for more illustrated details.

\textbf{Few-shots.} Without examples, LLM-generated SQL queries tend to be overly complex or inefficient. Providing well-designed few-shot examples helps the LLM produce more accurate and optimized queries by guiding it toward standard structures. These examples can be retrieved from a database, such as via an RAG pipeline.

\vspace{0.20 cm}

\textbf{Self-correction.} In the query generation process, errors often occur, requiring mechanisms like self-debugging or self-correction to make the output executable. These errors are typically syntax errors, where the SQL doesn’t follow proper syntax, and logical errors, where the query doesn’t capture the user’s intent. \citet{li2024codes} noted that self-correction is particularly effective for syntax errors, but has limited impact on logical inconsistencies. Based on this, our methodology uses self-correction as a fallback to fix any remaining syntax issues or incorrect data during the query cleaning process, ensuring the system’s reliability and that the resulting query answers the user’s problem accurately. The prompt for self-correction can be found at \textbf{Appendix \ref{selfcorrection}}.

\vspace{0.20 cm}

\textbf{Decomposition and Multistep generation.} To improve SQL generation quality, frameworks like MAC-SQL and E-SQL \cite{wang2023mac, caferouglu2024esql} break down queries into independent sub-queries, allowing the model to better understand the data structure, similar to an exploratory data analysis (EDA) phase, which reduces logical and data-related errors. However, this multi-step reasoning approach increases processing time due to the additional computational overhead from each sub-query.

\subsection{Metrics}
In text2sql systems, evaluation metrics typically fall into two categories \cite{mohammadjafari2024natural}:\textbf{ Content Matching-Based Metrics}, which include Component Matching (CM) and Exact Matching (EM), focusing on the alignment of SQL components or exact matches between generated and reference queries, and \textbf{Execution-Based Metrics}, such as Execution Accuracy (EX) and Valid Efficiency Score (VES), which assess the correctness and efficiency of query execution. 

However, these metrics have limitations in the financial domain: Content Matching ignores the correctness of query results, while Execution-Based Metrics fail to account for discrepancies due to extra or missing content in the output. Given the importance of correctness and completeness in financial data, we propose a hybrid evaluation approach that combines structural accuracy with contextual correctness and efficiency. This involves using an LLM as an evaluator to assess the quality of generated results against ground truth, by generating multiple-choice questions related to the original query, with the LLM selecting the correct answer or opting for "I don't know" if unsure. This method ensures a more reliable evaluation aligned with the real world requirements of financial statement analysis.

\section{Experiment}
\subsection{Setup}

\begin{table}[ht]
\centering
\resizebox{\columnwidth}{!}{%
\begin{tblr}{
  cell{2}{1} = {r=6}{},
  cell{8}{1} = {r=4}{},
  hline{1} = {-}{},
  hline{2,8,12} = {-}{},
}
\textbf{Category}             & \textbf{Model Name}                 & \textbf{Params} \\
\textbf{LLM}       & LLaMA 3.3 Instruct                  & 70B                     \\
                              & Qwen 2.5 Coder                      & 32B                     \\
                              & Gemini 2.0 Flash                    & -                       \\
                              & Gemini 2.0 Flash Think & -                       \\
                              & GPT-4o Mini                         & -                       \\
                              & Deepseek V3                         & 37B (671B)              \\
\textbf{SLM} & LLaMA 3.1 Instruct                  & 8B                      \\
                              & LLaMA 3.2 ~Instruct                 & 8B                      \\
                              & Qwen 2.5 Instruct                   & 3B – 7B                 \\
                              & Qwen 2.5 Coder Instruct         & 3B – 7B                 
\end{tblr}
}
\caption{Summary of Models Used in the Experiments.}
\label{tab:lms}
\end{table}
To evaluate our pipeline's effectiveness in processing Vietnamese financial statements, we tested a range of Language Models across two categories (see \textbf{Table \ref{tab:lms}}): Commercial LLMs and Small Language Models (SLMs). Where we denoted LLM: Commercial Large Language Model, SLM: Small Language Model. Deep Seek V3 having 671B params (37B activated each token), Qwen 2.5 has 2 model versions: 3B and 7B. Commercial LLMs are large, closed-source models accessed via API due to high resource demands, while smaller, open-source SLMs offer easier deployment and can achieve competitive performance with fine-tuning.

\begin{table}[ht]
\centering
\resizebox{\columnwidth}{!}{%
\begin{tblr}{
  hline{1} = {-}{},
  hline{2,13} = {-}{},
}
\textbf{Configuration}     & \textbf{Value/Details}                  \\
Training steps             & 8,000 steps (\textasciitilde{}4 epochs) \\
Batch size                 & 8                                       \\
Optimizer                  & AdamW                                   \\
Learning rate schedule     & Cosine decay \& warm-up  \\
LoRA rank                  & 64                                      \\
Hardware (1.5B  3B models) & NVIDIA RTX 4090                         \\
Hardware (7B model)        & NVIDIA A100                             \\
Training duration (7B)     & 16 hours                                    \\
Alignment Methods     & KTO \& DPO
\\
DPO Samples     & 2,000
\\
KTO Samples & 4,000
\end{tblr}}
\caption{Set up for training SLMs}
\label{tab:setups}
\end{table}

Further, we illustrate our setup and hyperparameters for SLM finetuning and techniques in \textbf{Table \ref{tab:setups}}. In our experiment, DPO (Direct Preference Optimization) \cite{rafailov2023dpo} and KTO (Kahneman-Tversky Optimization) \cite{ethayarajh2024kto} were used to align SLMs by leveraging GPT-4o-mini's incorrect outputs and Gemini's improved responses as training pairs, aiming to enhance reasoning in the text2sql pipeline.

In general, our \textbf{finstat2sql} is a fine-tuned version of Qwen 2.5 Coder for each model size, integrated with our proposed pipeline for enhanced performance.

\subsection{Result}
Proprietary models (\textbf{Table \ref{tab:ellm}}), particularly the Gemini family from Google, outperform all other models in the text2sql evaluation, with the "thinking" variant achieving the highest accuracy (72.03\%). Despite both Gemini 2.0 Flash and GPT-4o Mini being optimized for speed, Gemini demonstrates significantly better performance, suggesting its architecture or training data is better suited for this task. Among open-source models, DeepSeek-V3 performs competitively, likely due to its efficient MoE-based architecture using 37B active parameters. In contrast, dense models like Qwen2.5 32B Coder and LLaMA 3.3 70B Instruct show similar performance around 66\%, indicating a possible plateau at this scale. Overall, the evaluation highlights Gemini’s advantage in both speed and reasoning, affirmatively answering Research Question 2: the FinStat2SQL pipeline can meet over 72\% of real-world financial query needs when powered by Gemini’s reasoning model.

\begin{table}
\centering
\begin{tblr}{
  hline{1} = {-}{},
  hline{2,8} = {-}{},
}
\textbf{Model}            & \textbf{Params} & \textbf{Acc.} \\
gemini-2.0-flash-thinking & –               & \textbf{0.7203}   \\
\textit{gemini-2.0-flash} & –               & \textit{0.6969}   \\
deepseek-v3               & 37(671)B        & 0.6929            \\
qwen2.5-coder-instruct    & 32B             & 0.6590            \\
llama-3.3-instruct        & 70B             & 0.6569            \\
gpt-4o-mini               & –               & 0.5914            
\end{tblr}
\caption{Performance of LLMs on Evaluation set}
\label{tab:ellm}
\end{table}

\textbf{Table \ref{tab:eslm}} demonstrates the significant impact of task-specific fine-tuning on Small Language Models (SLMs) for the text2sql task. Across all sizes, the finstat2sql models (7B: 0.6133, 3B: 0.5558, 1.5B: 0.5301) consistently outperform their corresponding base models from the Qwen2.5 and Llama3 families, highlighting the benefits of specialization. While the Qwen2.5 coder models show stronger baseline performance than Llama3, fine-tuning proves to be the key differentiator most notably, the 7B finstat2sql model surpasses even the much larger 70B Llama and GPT-4o-mini in accuracy. This supports the conclusion that with targeted fine-tuning, smaller models can rival or even exceed the performance of much larger or proprietary alternatives, offering a cost-efficient solution for domain-specific tasks.

\begin{table}
\centering
\begin{tblr}{
  cell{2}{2} = {r=4}{},
  cell{6}{2} = {r=3}{},
  cell{9}{2} = {r=2}{},
  hline{1} = {-}{},
  hline{2,9,11} = {-}{},
  hline{6} = {1,3}{},
  hline{6} = {2}{},
}
\textbf{Model}            & \textbf{Params} & \textbf{Acc.} \\
qwen2.5-coder-instruct & 7–8B            & 0.4793            \\
qwen2.5-instruct       &                 & 0.4571            \\
llama3.1-instruct      &                 & 0.3804            \\
\textbf{finstat2sql}   &                 & \textbf{0.6133}   \\
qwen2.5-coder         & 3B              & 0.3019            \\
llama-3.2-instruct     &                 & 0.2126            \\
\textbf{finstat2sql}   &                 & \textbf{0.5558}   \\
qwen2.5-coder        & 1.5B            & 0.2299            \\
\textbf{finstat2sql} &                 & \textbf{0.5301}   
\end{tblr}
\caption{Evaluation on SLMs}
\label{tab:eslm}
\end{table}

\begin{table*}[ht]
\centering
\begin{tabular}{llll}
\hline
\textbf{Base model} & \textbf{Training type} & \textbf{Private test} & \textbf{Test Accuracy} \\ \hline
qwen2.5-coder-3b    & SFT                    & 0.7195                & \textbf{0.5558}        \\
                    & SFT+KTO                & \textbf{0.7257}       & 0.5204                 \\
                    & SFT+DPO                & 0.6761                & -                      \\ \hline
qwen2.5-coder-1.5b  & SFT                    & 0.6780                & \textbf{0.5301}        \\
                    & SFT+KTO                & \textbf{0.6879}       & 0.4419                 \\
                    & SFT+DPO                & 0.6288                & -                      \\ \hline
\end{tabular}
\caption{Evaluation on Model Alignment Methods}
\label{tab:rl}
\end{table*}

\textbf{Table \ref{tab:rl}} shows that alignment methods, especially DPO, significantly reduced performance, with both 3B and 1.5B models experiencing major accuracy drops on the private test set, hence further evaluation was skipped. While KTO slightly improved private test accuracy, it led to lower performance on the main evaluation set, suggesting overfitting and poor generalization.

\begin{figure}[h]
    \centering
    \includegraphics[width=\columnwidth, clip]{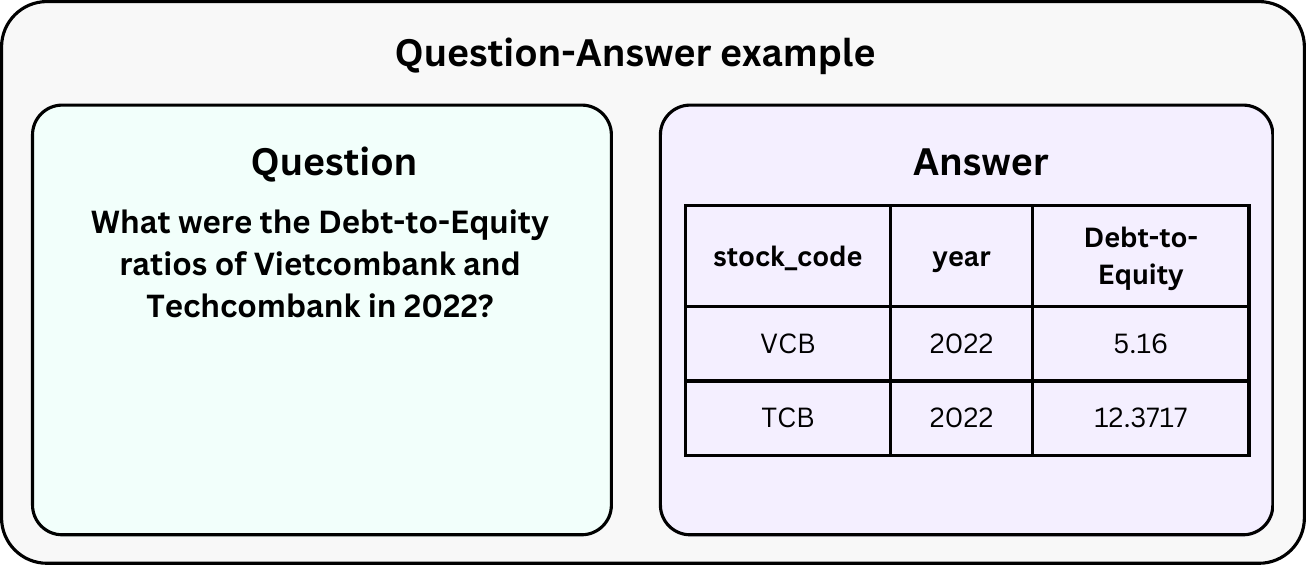}
    \caption{Question Answering with FinStat2SQL}
    \label{fig:qae}
\end{figure}

Finally, we demonstrate an example of integrating FinStat2SQL into a question-answering chatbot system. As shown in \textbf{Figure~\ref{fig:qae}}, once the user submits a query, the end-to-end pipeline returns the expected table. This table can then be further analyzed either by a human or by the chatbot for financial analysis.

\section{Conclusion}
\subsection{Discussion \& Conclusion}
This research demonstrates that FinStat2SQL allows users to query complex financial data using natural language, significantly lowering the barrier for non-experts. The pipeline balances accuracy and efficiency effectively, with simpler architectures often outperforming more complex ones. While proprietary models like Gemini-2.0-Flash-Thinking achieved the highest accuracy (72.03\%), open-source and fine-tuned models, especially DeepSeek V3 and Qwen2.5-Coder proved highly competitive. Notably, the 7B finstat2sql model matched or exceeded the performance of much larger models, highlighting the value of fine-tuned SLMs as cost-effective alternatives. Model alignment techniques were found to be unnecessary in many cases, and the system has been deployed as a financial chatbot with nearly 70\% query success and sub-4-second response times, addressing a critical gap in Vietnam’s financial automation landscape.

\subsection{Limitation \& Future Works}

This study has several limitations. First, the dataset is limited to VN30 and HNX30-listed firms, excluding Vietnam's small and medium enterprises (SMEs), which may limit generalizability. Second, the system struggles with Vietnamese financial terminology variations, potentially affecting entity recognition due to inconsistencies and biases toward international standards like IFRS over VAS. Third, FinStat2SQL has only been tested in the Vietnamese context and may not perform well with global frameworks like US GAAP. Lastly, deploying high-performing closed-source models such as GPT-4o-mini requires substantial computational resources, potentially restricting accessibility and real-time application due to cost and infrastructure constraints.

Future work will address these challenges by expanding the dataset to include SMEs, unlisted companies, and additional financial data sources for more comprehensive coverage. Efforts will also focus on adapting the pipeline for international standards to support cross-national applicability and integrating predictive analytics for trend forecasting and risk assessment. Given budget constraints, refining training methods and investigating why alignment techniques underperformed will be key, along with experimenting with data augmentation and improved fine-tuning strategies. These steps aim to enhance FinStat2SQL's robustness, utility, and scalability for both local and global financial analysis.

\bibliography{references}

\onecolumn
\appendix
\section{Appendix}
\subsection{Example of Evaluation Dataset}
\label{sec:appendixeval}

\textbf{Question: } Banks with credit growth higher than average in Q3 2023
\label{exgoldenquery}
\begin{tcolorbox}[
  colback=white, colframe=black, 
  width=\dimexpr\textwidth+3em\relax,
  boxrule=0.5pt,
  left=1em, right=1em,
  title=Golden Query,
  enlarge right by=3em
]
\begin{verbatim}
WITH bank_credit_growth AS (
 SELECT
 fr.stock_code,
 ci.industry,
 fr.year,
 fr.quarter,
 fr.data AS credit_growth_yoy
 FROM financial_ratio fr
 JOIN company_info ci ON fr.stock_code = ci.stock_code
 WHERE fr.ratio_code = 'CDGYoY'
 AND ci.is_bank = TRUE
 AND fr.year = 2023
 AND fr.quarter = 3
),
industry_avg_growth AS (
 SELECT
 industry,
 year,
 quarter,
 data_mean AS industry_credit_growth
 FROM industry_financial_ratio
 WHERE ratio_code = 'CDGYoY'
 AND industry = 'Banking'
 AND year = 2023
 AND quarter = 3
)
SELECT
 b.stock_code,
 b.year,
 b.quarter,
 b.credit_growth_yoy,
 i.industry_credit_growth
FROM bank_credit_growth b
JOIN industry_avg_growth i ON b.industry = i.industry
WHERE b.credit_growth_yoy > i.industry_credit_growth
ORDER BY b.credit_growth_yoy DESC
\end{verbatim}
\end{tcolorbox}

\begin{table}[h]
\centering
\caption{Example Golden Query result table}
`\begin{tabular}{lllll}
\hline
\textbf{Stock code} & \textbf{Year} & \textbf{ Quarter} & \textbf{ Credit-YoY} & \textbf{Industry YoY} \\
\hline
HDB        & 2023 & 3       & 0.64       & 0.24         \\
VPB        & 2023 & 3       & 0.52       & 0.24         \\
MSB        & 2023 & 3       & 0.35       & 0.24         \\
KLB        & 2023 & 3       & 0.34       & 0.24         \\
\ldots     & \ldots & \ldots & \ldots    & \ldots       \\
\hline
\end{tabular}
\end{table}

\subsection{Prompt Detail}
\subsubsection{Entity extraction and Row selection Agent Prompt}
\label{entityprompt}
\begin{tcolorbox}[
  colback=white, colframe=black, 
  width=\dimexpr\textwidth+3em\relax,
  boxrule=0.5pt,
  left=1em, right=1em,
  title=User Prompt,
  enlarge right by=3em,
]
\begin{Verbatim}[]
<task>
Based on given question, analyze and suggest the suitable accounts in the financial
statement and/or financial ratios that can be used to answer the question.
Extract the company name and/or the industry that positively mentioned based on the
given question.
</task>
<question>
{task}
</question>
Analyze and return the accounts and entities that useful to answer the question.
Return in JSON format, followed by this schema.
```
{{
 "industry": list[str],
 "company_name": list[str],
 "financial_statement_account": list[str],
 "financial_ratio": list[str]
}}

```
Note:
- Return an empty list if no related data is found.
- If the question include tag "4 nearest quarter", "trailing twelve months
(TTM)","YoY" or "QoQ", include the name and tag into `financial_ratio` and/or
`financial_statement_account`
- If "YoY" and "QoQ" ratio are mentioned, include the original account in
`financial_statement_account` and the ratio in `financial_ratio`.
<example>
### Question:
Net Income YoY and ROE 4 nearest quarter of HPG in 2023
### Response:
```json
 {{
 "industry": [],
 "company_name": ["HPG"],
 "financial_statement_account": ["Net Income"],
 "financial_ratio": ["Net Income YoY", "ROE 4 nearest quarter"]
 }}
```
</example>
\end{Verbatim}
\end{tcolorbox}
\vspace{1em}

\subsubsection{Schema Description Prompt}
\label{schemadesc}
\begin{tcolorbox}[
  colback=white, colframe=black, 
  width=\dimexpr\textwidth+3em\relax,
  boxrule=0.5pt,
  left=1em, right=1em,
  title=System prompt,
  enlarge right by=3em,
]
\begin{Verbatim}
<overall_description>
The database conatains financial statments of Vietnamese firms, followed by the
regulation of Vietnamese Accounting Standard (VAS). The database includes two
reporting periods: quarterly (1, 2, 3, 4) and annually (quarter = 0).
</overall_description>
### PostgreSQL tables in OpenAI Template
<schema>
- **company_info**(stock_code, industry, exchange, stock_indices, is_bank,
is_securities)
- **sub_and_shareholder**(stock_code, invest_on)
- **financial_statement**(stock_code, year, quarter, category_code, data,
date_added)
- **industry_financial_statement**(industry, year, quarter, category_code,
data_mean, data_sun, date_added)
- **financial_ratio**(ratio_code, stock_code, year, quarter, data, date_added)
- **industry_financial_ratio**(industry, ratio_code, year, quarter, data_mean,
date_added)
- **financial_statement_explaination**(category_code, stock_code, year, quarter,
data, date_added)
</schema>
### Note on schema description:
- For industry tables, column `data_mean` is average data of all firms in that 
industry, while `data_sum` is the sum of them.
- Table `financial_statement_explaination` contains information which is not covered
in 3 main reports, usually about type of loans, debt, cash, investments and 
real-estate ownerships.
- With YoY ratio in `financial_ratio`, you should recalculate the ratio if the time
window is not 1 year.
### Note on query:
- You will be provided a mapping table for `category_code` and `ratio_code` to select
suitable code.
- For any financial ratio, it must be selected from the database rather than being
calculated manually.
- Always include a `quarter` condition in your query. If not specified, assume using
annual reports (`quarter` = 0).
- Always include LIMIT.
\end{Verbatim}
\end{tcolorbox}

\vspace{1em}
\label{selfcorrection}
\subsubsection{Self Correction Prompt}
\begin{tcolorbox}[
  colback=white, colframe=black, 
  width=\dimexpr\textwidth+3em\relax,
  boxrule=0.5pt,
  left=1em, right=1em,
  title=User Prompt,
  enlarge right by=3em,
]
\begin{Verbatim}
<result>
{sql_result}
</result>

<correction>
Based on the SQL table result in <result> tag, do you think the SQL queries is correct
and can fully answer the original task? If there is no SQL Result table on <result> 
tag, it means the preivous queries return nothing, which is incorrect.
If the result of SQL query is correct and the table is suitable for <task> request,
you only need to return YES under *Decision* heading. You must not provide the SQL
query again.
Otherwise, return No under *Decision* heading, think step-by-step under
*Reasoning* heading again and generate the correct SQL query under *SQL Query*.
Return in the following format (### SQL Query is optional):
### Decision:
{{Your decision}}
### Reasoning:
{{Your reasoning}}
### SQL Query:
{{Corrected SQL query}}
</correction>
\end{Verbatim}
\end{tcolorbox}

\subsection{Database Design}
\label{erd}
\begin{figure*}[h]
    \centering
    \includegraphics[width=\textwidth, clip]{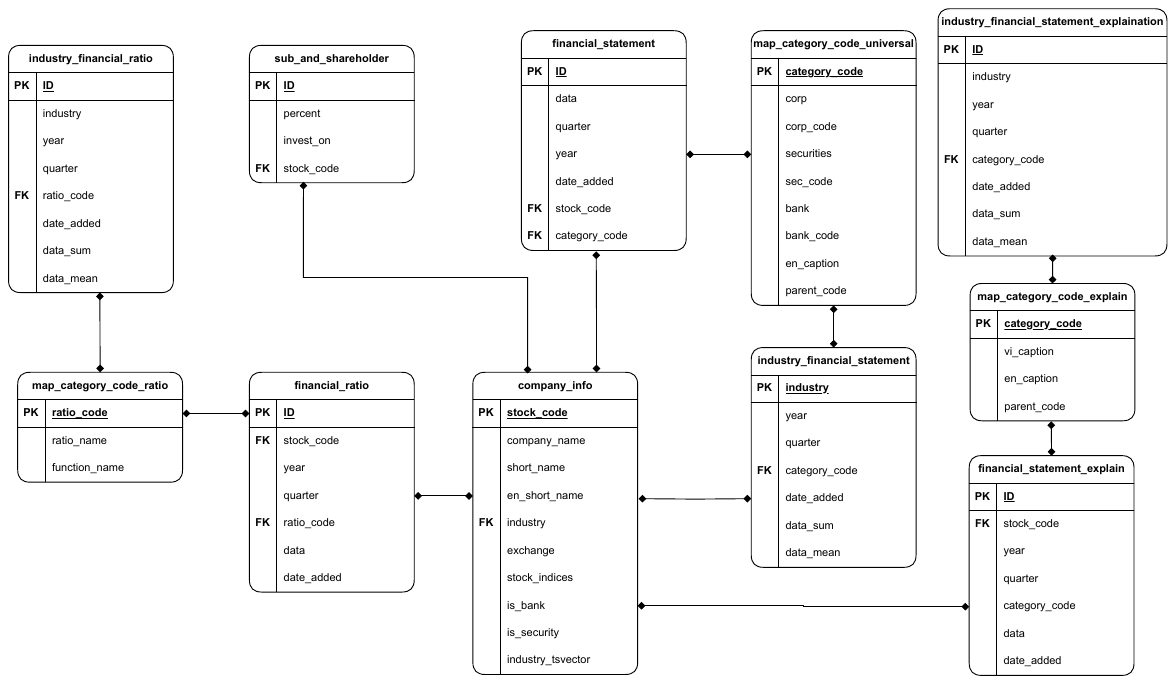}
\end{figure*}

\end{document}